\documentclass[12pt]{article}

\usepackage[margin=1in]{geometry}
\usepackage[T1]{fontenc}
\usepackage[utf8]{inputenc}
\usepackage{lmodern}
\usepackage{microtype}
\usepackage{amsmath,amssymb}
\usepackage{booktabs}
\usepackage{xcolor}
\usepackage{fancybox}
\usepackage{graphicx}
\usepackage{authblk}
\usepackage{hyperref}
\hypersetup{colorlinks=true, linkcolor=blue, citecolor=blue, urlcolor=blue}

\title{\textbf{Anonymous-by-Construction: An LLM-Driven Framework\\
for Privacy-Preserving Text}}

\author[1,2]{Federico Albanese\thanks{Corresponding author: \texttt{falbanese@dc.uba.ar}}}
\author[1]{Pablo Ronco}
\author[1,3]{Nicol\'{a}s D'Ippolito}

\affil[1]{Veritran, Buenos Aires, Argentina}
\affil[2]{University of Buenos Aires, Argentina}
\affil[3]{University of San Andr\'{e}s, Buenos Aires, Argentina}

\date{2026}

\begin{document}

\maketitle

\begin{abstract}
Responsible use of AI demands that we protect sensitive information without undermining
the usefulness of data, an imperative that has become acute in the age of large language
models. We address this challenge with an on-premise, LLM-driven substitution pipeline
that anonymizes text by replacing personally identifiable information (PII) with
realistic, type-consistent surrogates. Executed entirely within organizational
boundaries using local LLMs, the approach prevents data egress while preserving fluency
and task-relevant semantics.

We conduct a systematic, multi-metric, cross-technique evaluation on the Action-Based
Conversation Dataset, benchmarking against industry standards (Microsoft Presidio and
Google DLP) and a state-of-the-art approach (ZSTS, in redaction-only and
redaction-plus-substitution variants). Our protocol jointly measures privacy (PII
recall), semantic utility (sentiment agreement, topic distance), and trainability under
privacy via a lifecycle-ready criterion obtained by fine-tuning a compact encoder
(BERT+LoRA) on sanitized text. In addition, we assess agentic Q\&A performance by
inserting an on-premise anonymization layer before the answering LLM and evaluating the
quality of its responses. This intermediate, type-preserving substitution stage ensures
that no sensitive content is exposed to third-party APIs, enabling responsible deployment
of Q\&A agents without compromising confidentiality.

Our method attains state-of-the-art privacy, minimal topical drift, strong factual
utility, and low trainability loss, outperforming rule-based approaches and named-entity
recognition (NER) baselines and ZSTS variants on the combined
privacy--utility--trainability frontier. These results show that local LLM substitution
yields anonymized corpora that are both responsible to use and operationally valuable:
safe for agentic pipelines over conversational data and suitable for downstream
fine-tuning with limited degradation. By coupling on-premise anonymization with rigorous,
auditable evaluation, this work offers decision-ready evidence for deploying
Responsible-AI-by-Design systems in real-world settings.
\end{abstract}


\bigskip

\section{Introduction}

The accelerating integration of AI, particularly LLM-centric compound systems and
emerging agentic workflows, has amplified long-standing concerns about the handling of
sensitive information. Engineering teams are expected to deliver features that learn from
user interactions, interact with third-party tools, and evolve continuously, while also
complying with privacy and safety regulations such as the General Data Protection
Regulation (GDPR)~\cite{goddard2017eu}, the Health Insurance Portability and
Accountability Act (HIPAA)~\cite{nosowsky2006health} and the California Consumer Privacy
Act (CCPA)~\cite{goldman2020introduction} in the US and EU regions. In parallel,
organizations are increasingly expected to align with ISO/IEC
42001~\cite{dudley2024rise}, which emphasizes data governance and confidentiality
throughout the AI lifecycle---complementing information security controls to ensure that
sensitive user information remains protected by design. This is especially critical in
highly regulated sectors such as healthcare, financial services, and banking, where data
breaches or misuse can trigger severe regulatory exposure, fiduciary risks, and direct
harm to individuals.

From a Responsible-AI Engineering perspective, this tension surfaces throughout the
system lifecycle: requirements (privacy and utility as quality attributes), architecture
(data boundaries, on-prem isolation), verification \& validation (privacy/utility gates),
and operations (LLMOps/AgentOps with auditability, rollback, and
monitoring)~\cite{lu2024responsible, xia2024towards, bi2025privacy, dong2024agentops}.
A persistent, practical gap remains: How can we produce anonymous yet useful data that
(i) are safe to process with third-party APIs, LLMs, and agents, (ii) preserve downstream
utility, and (iii) support fine-tuning without incentivizing memorization or enabling
extraction attacks?

We address this challenge with an on-premise, LLM-driven substitution anonymization
pipeline. Rather than deleting spans (i.e., contiguous text segments) through
redaction-only, it rewrites each sentence in place, replacing sensitive content with
realistic, type-preserving surrogates. The method executes strictly within organizational
boundaries using local runtimes
(e.g., GPT-oss 20B~\cite{openai2025gptoss120bgptoss20bmodel},
DeepSeek-r1 7B~\cite{deepseekai2025deepseekr1incentivizingreasoningcapability}),
preventing data egress and aligning with privacy-by-design and safety-aware architectural
principles~\cite{lu2024responsible}.

While recent work has explored LLM anonymizers~\cite{wiest2025deidentifying,
staab2025language}, our paper instead provides a systematic, multi-metric, cross-technique
evaluation that turns fragmented evidence into operational guidance. Specifically, we
jointly assess privacy, semantic and factual utility, agentic Q\&A performance, and
trainability under privacy (LoRA fine-tuning) across multiple anonymization methods:
Google DLP~\cite{Google}, Microsoft Presidio~\cite{MSFT}, Bert zero-shot text
sanitization (redaction and redaction+substitution)~\cite{albanese2023text}, and
LLM-based substitution; under a unified protocol.

The main contribution of this work is an anonymization framework designed to enforce
privacy guarantees, with the following characteristics:
\begin{itemize}
  \item \textbf{Local processing.} All operations execute on organizational infrastructure
    (e.g., self-hosted LLM runtimes) without exposing data to third-party services,
    eliminating egress risks and satisfying regulatory privacy requirements.
  \item \textbf{Minimal adaptation.} The method uses few-shot prompting without
    task-specific training, providing domain independence and enabling immediate
    deployment in production settings.
  \item \textbf{Language agnostic.} By leveraging pre-trained multilingual LLMs, the
    framework processes text across languages without retraining, supporting heterogeneous
    datasets through a single pipeline.
  \item \textbf{Downstream fidelity.} Contextual substitution preserves \emph{Sentiment}
    labels, \emph{Topic} representations, and factual accuracy in \emph{Q\&A}
    evaluations, while maintaining model trainability on sanitized corpora.
\end{itemize}

In sum, the proposed framework provides foundational support for Responsible-AI pipelines:
agentic systems operate on anony\-mous-by-construction data; model fine-tuning proceeds
without memorization risks; and privacy is enforced as a first-class design constraint
throughout the lifecycle. Crucially, by measuring privacy, utility, and trainability
together, our study offers decision-ready evidence for LLMOps / AgentOps teams tasked
with deploying trustworthy AI over sensitive data.

The remainder of the paper is organized as follows: Section~2 details the LLM-Driven
Substitution Anonymization method on site, including prompt policy, execution environment,
and RAIE controls. Section~3 (Evaluation) specifies the data set, baselines and evaluation
framework (privacy preservation, sentiment, topic, Q\&A, and LoRA-based fine-tuning).
Section~4 (Results) presents the comparative findings and analyzes privacy--utility
trade-offs. Section~5 (Related Work) situates our approach within prior text sanitization
and responsible-AI engineering literature. Section~6 (Conclusions) summarizes
contributions, discusses limitations, and outlines future work.

\section{On-Premise LLM-Driven Substitution Anonymization}

We anonymize conversational text via prompted, type-consistent substitution using
on-premise large language models (LLMs). Rather than deleting spans (redaction-only),
the model rewrites each utterance in place, replacing names, usernames, emails, phone
numbers, addresses, codes, and all numeric expressions (including dates) with realistic
but fictitious alternatives of the same type. This preserves fluency and task-relevant
semantics while removing personally identifiable information (PII).

In this paper, we run two open-source local LLMs: GPT-oss
(20B)~\cite{openai2025gptoss120bgptoss20bmodel}, and DeepSeek-r1
(7B)~\cite{deepseekai2025deepseekr1incentivizingreasoningcapability}.

At the core of the method is a single instruction that enforces (i)~span detection by
type, (ii)~type-preserving replacement, and (iii)~faithful reproduction of the original
structure. The instruction is issued utterance-by-utterance to emulate a streaming chat
backend. The utterance is encoded into a prompt, whose template is shown in
Figure~\ref{fig:prompt}.

\begin{figure}[t]
\centering
\shadowbox{%
\begin{minipage}{0.85\textwidth}
\textbf{Prompt}

\vspace{0.6em}

Redact and replace the words in the following text that correspond to numbers and nouns,
such as names, surnames, numbers, usernames, codenames, emails, phone numbers, addresses,
or other data. Replace all numbers, including dates, codes, etc.\ with other realistic but
fake alternatives of the same type. Go word by word in the original text and replace the
words with realistic and similar alternatives of the same type. For example, if the text
contains a name like `John Doe', replace it with `David Smith'. For example, if the text
contains an email like `johndoe@gmail.com', replace it with `davidsmith@hotmail.com'. For
example, if the text contains a number like `762991, 1st', replace it with a number of the
same length `951910, 2nd'. Do not eliminate any sentence or part of the text, just replace
the words with alternatives. Return only the modified text without any additional text or
explanations. Text:\textit{\{original\_text\}}

\end{minipage}
}
\caption{Prompt template used for text anonymization. The instruction enforces type-aware
  detection and replacement of personally identifiable information (PII), ensuring that
  each sensitive span is substituted with a realistic but fictitious instance of the same
  semantic type while preserving the original text structure and fluency.}
\label{fig:prompt}
\end{figure}

We use low-stochasticity decoding (i.e., zero temperature) to minimize variance across
runs and ensure deterministic replacements under a fixed seed.

Both GPT-oss (20B) and DeepSeek-r1 (7B) operate on local infrastructure without external
data transfer, satisfying organizational privacy constraints while enabling comparison
between model scale and computational cost.

Prompted substitution with local LLMs thus provides a practical, auditable path to
privacy-preserving data pipelines that operates entirely within organizational boundaries,
without exposing sensitive information to external APIs or services.

\section{Evaluation}

\subsection{Datasets}

We conduct our evaluation on the Action-Based Conversation Dataset
(ABCD)~\cite{chen2021action}, a widely used benchmark for task-oriented dialogue. ABCD
comprises more than $10{,}000$ human-to-human conversations with $55$ user intents and
over $140{,}000$ utterances. Notably, this corpus contains metadata identifying words
corresponding to (synthetic) personally identifiable information (PII).

We consider the following metadata fields as PII: customer name, username, email, phone
number, account ID, order ID, street address, and ZIP code. These categories align with
the dataset's slot annotations, enabling systematic evaluation of the privacy-utility
trade-off.

\subsection{Baselines}

To contextualize our approach, we benchmark against two widely used de-identification
systems that represent the prevailing industrial practice, and one recent state-of-the-art
method tailored to text sanitization:

\begin{itemize}
  \item \textbf{Microsoft Presidio~\cite{MSFT}.} Presidio is an open-source PII
    sanitization toolkit that couples pattern rules with NER-based recognizers
    (e.g., spaCy/Stanza or custom models), enriched with context (``hotword'') signals
    and confidence scoring. Spans flagged as sensitive can be masked, replaced via
    templates, hashed, or tokenized with format preservation. Thanks to its decoupled
    Analyzer/Anonymizer architecture, extensible recognizer ecosystem, and deterministic
    transformation operators, Presidio fits pipelines that demand auditability,
    repeatability, and policy-governed controls.

  \item \textbf{Google Data Loss Prevention (DLP)~\cite{Google}.} DLP is a managed
    inspection and de-identification service that detects a broad catalog of infoTypes
    using pattern matching, checksums, machine-learned detectors, and context rules
    (e.g., proximity constraints). It offers character masking, generalization,
    redaction, deterministic tokenization, and format-preserving encryption, all
    configurable through declarative policies with likelihood thresholds and rule
    overrides.

  \item \textbf{Zero-Shot Redaction and Substitution with Large Language Models
    (ZSTS)~\cite{albanese2023text}.} ZSTS frames privacy as detecting low-likelihood
    tokens under a masked language model and then redacts or substitutes them to balance
    privacy and semantic retention. Concretely, a masked LLM (in our case, BERT) scores
    tokens; those below a privacy threshold $p$ are candidates for removal/replacement.
    Substitutions are drawn from top-$n$ language-model proposals and filtered using
    embedding similarity within a radius $s$, with optional randomness over the top-$k$
    to hinder reverse mapping. ZSTS has reported strong performance on ABCD. We,
    therefore, include it as a state-of-the-art comparator. For fair evaluation, we
    replicate the original configuration: $\mathit{LLM}=\mathrm{BERT}$,
    $p=1\times10^{-1}$, $n=50$, $s=2$, $k=1$). We report results for both ZSTS
    configurations: Redaction-only and Redaction+Substitution.
\end{itemize}

We adopt Google DLP and Microsoft Presidio as comparative baselines because they are
widely deployed and frequently cited in prior work, providing a reproducible,
policy-oriented reference point for de-identification
research~\cite{hassan2021utility, lison2021anonymisation, zhou2023privacy}. In addition,
we also include ZSTS as a recent state-of-the-art text sanitization method.

\subsection{Experiments}

We now describe our evaluation framework, designed to quantify the privacy--utility
trade-off under realistic deployment conditions. Unless explicitly stated otherwise, all
experiments are conducted on the complete set of ABCD conversations.

\subsubsection{Privacy Preservation}

We frame privacy preservation as detecting unsafe (PII) spans and removing or substituting
them. Accordingly, we adopt a standard classification metric: recall.

\begin{displaymath}
  \mathit{Recall} =
  \frac{\mathit{TruePositives}}{\mathit{TruePositives} + \mathit{FalseNegatives}}
\end{displaymath}

\noindent where:
\begin{itemize}
  \item $\mathit{TruePositives}$ is the number of unsafe terms properly predicted as
    unsafe; and
  \item $\mathit{FalseNegatives}$ is the number of unsafe terms wrongly predicted as safe.
\end{itemize}

The recall measures the fraction of ground-truth unsafe terms that are successfully
sanitized. High recall implies that the anonymi\-zer eliminates most sensitive mentions,
yielding text that is effectively free of personal information and therefore ready for use
in downstream AI pipelines responsibly. On the other hand, low recall indicates that many
unsafe spans remain and the processed text would not constitute responsible use because it
still carries sensible information.

Using ABCD's slot annotations as the reference set of PII, we automatically count the
sensitive values that survive post-processing and report Privacy [Recall] in our results
table.

\subsubsection{Sentiment Analysis}

To evaluate whether sanitization alters affective content, we compare the sentiment label
of each text before and after anonymization. Following prior work, we use the off-the-shelf
VADER scorer~\cite{hutto2014vader} to assign each utterance (or conversation segment) a
polarity between $-1$ and $1$ and map it to one of the following: positive, neutral, or
negative. In VADER, scores $>0.05$ are labeled positive, scores $<-0.05$ are negative, and
values in between are neutral.

We then compute Sentiment Accuracy as the fraction of instances where the post-sanitization
label matches the original. This provides a task-level indicator of whether processing the
text preserves pragmatic cues such as tone and stance.

\subsubsection{Topic Embedding}

We also evaluate topical stability with an unsupervised topic embedding approach:
BERTopic~\cite{grootendorst2022bertopic}. Each conversation is projected into a
topic-probability vector using a transformer encoder plus clustering. We then compute the
cosine distance between the topic vectors of the original and sanitized texts. Smaller
values indicate better topical preservation, whereas larger values indicate that the topic
representation changed significantly. BERTopic finds $33$ topics in ABCD.

The results table reports Topic distance $[\text{mean} \pm \text{std}]$. This metric
captures whether the anonymization pipeline maintains the conversation's subject matter
even when tokens are removed or replaced.

\subsubsection{Question Answering (Q\&A)}

We evaluate the effect of sanitization on factual utility using GPT-3.5~\cite{brown2020language}.
The test suite targets:
\begin{itemize}
  \item entity-centric queries (names, usernames, emails)
  \item signs of dissatisfaction from the interlocutors
  \item whether the conversation is complete or was interrupted.
\end{itemize}

For each item, we issue the same prompt to both the raw and the sanitized conversation.
When a method performs substitution, we also apply the inverse mapping from the
substitution table at evaluation time to reconstruct original surface forms where needed
for an apples-to-apples comparison, following standards in the
literature~\cite{albanese2023text}.

We score responses with Accuracy, assigning 1 if the answer on the sanitized text matches
the answer on the original (unredacted) conversation, and 0 otherwise. Because
conversations without entities make it trivially easy to report that ``no entity is
present,'' we additionally report Q\&A true---accuracy restricted to cases where the
original answer is verified correct and requires recovering or preserving an entity mention.
In accordance with previous work~\cite{albanese2023text}, we run this protocol on a random
subset of 50 ABCD conversations.

This evaluation is particularly important for agentic pipelines that rely on third-party
LLMs or external APIs for automatic entity extraction and for computing conversation-level
KPIs (e.g., who was mentioned, whether a complaint was expressed, or whether a ticket
remains unresolved). Our goal is to ensure that the processed text remains
anonymous---the agent is blind to sensitive information---while still enabling the system
to respond accurately and identify task-relevant facts. In short, responsible AI agents
should preserve privacy by construction without sacrificing the operational utility
required for downstream analytics and decision-making.

\subsubsection{Fine-Tuning LLMs}

To assess whether sanitized text remains trainable for downstream modeling, we fine-tune a
compact encoder on a regression proxy for utility. In particular, we use
\textit{bert-base-uncased} from HuggingFace~\cite{DBLP:journals/corr/abs-1810-04805} and
Low-Rank Adaptation (LoRA)~\cite{hu2022lora}, a technique for efficiently fine-tuning large
models by adding small, trainable low-rank matrices to the original model. We configure LoRA
with the following parameters: $r=8$, $\mathit{lora\_alpha}=32$, and $\mathit{dropout}=0.1$.
The data are split $80/20$ into train and test. The prediction task is to estimate the VADER
sentiment score computed on the original (unsanitized) conversation using only the sanitized
conversation as input; the original text is never used as model input. We report the Mean
Absolute Error (MAE) on the held-out test set. A lower MAE indicates that sanitization
preserved the signal necessary for effective fine-tuning and a higher downstream utility
under privacy constraints.

This metric is central to a Responsible AI pipeline: training on sensitive text is
inappropriate because information seen during training can later be recovered by adversarial
prompts, exposing personally identifiable information~\cite{brokman2025insights}. Prior
studies have reported such memorization and extraction risks when models ingest sensitive
content~\cite{he2025security}. Our protocol therefore sanitizes first and then measures how
much predictive performance is lost when learning from sanitized inputs only, using the
original sentiment signal as a label. Ideally, the gap is as small as possible, indicating
that the anonymization process has maintained the intended balance: it removes or substitutes
PII while preserving the task-relevant structure needed for model performance. This metric
directly supports privacy-by-design and safety-by-default by aligning model development
decisions with quantifiable, pre-declared privacy and utility
thresholds~\cite{xia2024towards}.

\section{Results}

Table~\ref{tab:privacy_utility_trainability} summarizes the comparative results across all
baselines and our approach. The table aggregates privacy, semantic utility, factual utility
in agentic Q\&A, and downstream fine-tuning viability under privacy constraints.

Table~\ref{tab:privacy_utility_trainability} reports privacy, utility, and trainability
metrics for all baselines and our approach. We evaluate Privacy as redaction Recall over
ABCD's PII slots; Sentiment as label Accuracy pre/post sanitization; Topic distance as the
$\text{mean}\pm\text{std}$ cosine distance between topic vectors of original vs.\ sanitized
conversations; Q\&A / Q\&A true as factual Accuracy (overall vs.\ entity-dependent subset);
and fine-tuning as MAE using BERT+LoRA.

\begin{table}[t]
\caption{Privacy--Utility--Trainability comparison on ABCD. Privacy is redaction Recall;
  Sentiment is label Accuracy pre/post sanitization; Topic distance is cosine distance
  ($\text{mean}\pm\text{std}$) between topic vectors; Q\&A and Q\&A true are factual
  Accuracies (overall vs.\ entity-dependent subset); LoRA fine-tuning is MAE when
  predicting original VADER scores from sanitized text with BERT+LoRA.}
\label{tab:privacy_utility_trainability}
\centering
\small
\resizebox{\textwidth}{!}{%
\begin{tabular}{lcccccc}
\toprule
\textbf{Method}
  & \textbf{Privacy}
  & \textbf{Sentiment}
  & \textbf{Topic dist.}
  & \textbf{Q\&A}
  & \textbf{Q\&A true}
  & \textbf{LoRA fine-tuning} \\
  & \textbf{[Recall]}
  & \textbf{[Accuracy]}
  & \textbf{[mean$\pm$std]}
  & \textbf{[Accuracy]}
  & \textbf{[Accuracy]}
  & \textbf{[MAE]} \\
\midrule
Presidio (Microsoft)           & 0.56 & 0.999 & $0.001 \pm 0.001$   & 0.61 & 0.52 & 0.030 \\
DLP (Google)                   & 0.65 & 1.000 & $0.0003 \pm 0.0006$ & 0.52 & 0.36 & 0.032 \\
\midrule
ZSTS Redaction                 & 0.98 & 0.997 & $0.022 \pm 0.021$   & 0.57 & 0.26 & 0.417 \\
ZSTS Redaction \& Substitution & 0.98 & 0.993 & $0.023 \pm 0.021$   & 0.82 & 0.75 & 0.055 \\
\midrule
LLM (GPT-oss)                  & 0.99 & 1.000 & $0.002 \pm 0.001$   & 0.95 & 0.96 & 0.029 \\
LLM (DeepSeek-r1)              & 0.99 & 1.000 & $0.003 \pm 0.001$   & 0.93 & 0.93 & 0.038 \\
\bottomrule
\end{tabular}%
}
\end{table}

Classical baselines such as Microsoft Presidio and Google DLP achieve near-ceiling
Sentiment accuracy and negligible Topic distance, indicating minimal semantic drift.
However, their Privacy recall is low ($0.56$ and $0.65$ respectively). This implies that
substantial portions of annotated PII remain unsanitized, limiting their suitability for
responsible training or sharing with an LLM. Their fine-tuning MAE is competitive with
the other models. A possible explanation for this result is the limited perturbation of
the original text. In other words, it fails to remove all sensitive information. In
short, the downstream scores are good partly because the text is left largely intact,
which violates responsible use requirements.

ZSTS Redaction pushes privacy recall to $0.98$ but severely harms Q\&A true accuracy
($0.26$) and, critically, yields a fine-tuning MAE of $0.417$, indicating that the
resulting corpus is poorly suited for downstream tasks. This occurs because the method
only deletes tokens; achieving $0.98$ privacy requires removing many words, which harms
utility. However, adding Substitution substantially improves Q\&A / Q\&A true ($0.82$
and $0.75$ respectively) and reduces fine-tuning MAE to $0.055$ by reinjecting
semantically compatible substitutes, yet it still falls short of our approach on both
factual utility and trainability. Moreover, a Topic distance of $0.023$ further suggests
noticeable topical distortion relative to other methods.

Finally, our approach (the last two rows) delivers state-of-the-art privacy recall
($0.99$) while maintaining high sentiment accuracy and low topic drift ($0.002$ and
$0.003$). Crucially, factual utility is preserved: Q\&A / Q\&A true reach $0.95$ and
$0.96$ for GPT-oss and $0.93$ and $0.93$ for DeepSeek-r1, outperforming all baselines,
including ZSTS with substitution.

From a Responsible-AI engineering standpoint, the decisive result is fine-tuning MAE:
$0.029$ (GPT-oss) and $0.038$ (DeepSeek-r1). This demonstrates that our
substitution-based anonymization produces corpora that are safe-to-fine-tune while
retaining the signal needed for downstream adaptation.

Methods that redact less, like Presidio or DLP, unsurprisingly score well on downstream
metrics but underperform on privacy, making them inappropriate for training or release.
Pure redaction (ZSTS) maximizes privacy at the cost of utility and trainability.

Our approach yields privacy-preserving text that remains operationally valuable: it is
anonymized enough to be responsibly used and trainable enough to support downstream
learning with minimal degradation, precisely the goal of a Responsible-AI pipeline. In
particular, the strong Q\&A results make our approach attractive for agentic frameworks
that apply third-party LLMs to chat-based conversations (e.g., automatic entity
extraction and KPI computation) while keeping agents blind to sensitive data. At the same
time, the low fine-tuning MAE enables fine-tuning on sanitized corpora without
compromising privacy. Together, these properties support a Responsible-AI-by-Design and
AI-Safety-by-Design software architecture for handling sensitive conversational data.

\section{Related Work}

Automated anonymization of textual data has received sustained attention.
Surveys~\cite{lison2021anonymisation} map the main concepts and families of methods;
however, much of the prior work relies on rule-based approaches and named-entity
recognition (NER), which are limited to predefined sets of rules and entity types. For
instance, \cite{davidson2021improved} presents a NER-based pipeline tailored to identify
spans requiring redaction, while \cite{mehta2019towards} employs a two-stage Conditional
Random Field (CRF) model to perform NER over unstructured text for anonymization.
However, NER methods are constrained by a fixed ontology of entity types and rely on
supervised classifiers that demand substantial manually annotated corpora to be
trained~\cite{hassan2021utility}. In contrast, our method leverages a pretrained LLM,
delivering faster time-to-value in production pipelines and remaining applicable across
diverse contexts precisely because it does not rely on predefined entity inventories.

A closely related line of work adopts language-model-guided sanitization. For example,
\cite{papadopoulou2022neural} protects privacy by evaluating and masking both direct and
quasi-identifiers in text, using a masked LM (e.g., BERT) to estimate term probabilities
as a proxy for sensitivity. In a similar spirit, ZSTS~\cite{albanese2023text} leverages
a masked LM to score tokens against a privacy threshold and then performs redaction and,
optionally, semantics-preserving substitution to balance privacy with utility. Both
approaches thus employ pretrained LMs (notably BERT) to quantify token-level risk.

Building on these LM-guided schemes that quantify token-level risk, subsequent research
has examined how to replace rather than merely mask sensitive spans to better preserve
utility. Prior work has often treated detected spans with template-based redaction,
replacing them with default strings or placeholders~\cite{braathen2021creating,
papadopoulou2022neural}. In contrast, a growing body of work emphasizes
semantics-preserving substitution to maintain contextual meaning and discourse coherence.
For example, \cite{sanchez2013automatic} proposes a general-purpose approach that
leverages knowledge bases and term-frequency statistics to choose substitutes for
sensitive terms; and \cite{olstad2023generation} generates replacements via a combination
of heuristic rules and an ontology derived from Wikidata. Complementing these
resource-driven strategies, ZSTS~\cite{albanese2023text} selects substitutes by scoring
tokens with a masked language model (BERT) and filtering candidates using
embedding-similarity constraints, thereby balancing privacy with semantic fidelity.

Recent work has explored LLM-based anonymization with complementary but partial scopes.
First, \cite{wiest2025deidentifying} employs an LLM to detect sensitive mentions in
clinical text but does not perform substitution, losing text utility. Second,
\cite{staab2025language} advances to detection-and-replacement with an LLM, yet provides
only limited evaluation of downstream utility for the substituted outputs. Closer to our
setting, \cite{yang2025gama} performs detection and substitution and assesses utility
with Q\&A agents, underscoring that semantic preservation is pivotal for agentic
pipelines. In a complementary direction, \cite{gardiner2024data} examines whether
fine-tuning on anonymized data preserves performance---finding that it does---but
constrains anonymization to Google DLP (one of our baselines) and does not compare
against alternative techniques or introduce an operational criterion for trainability.
Against this backdrop, our contribution fills a crucial gap by delivering a systematic,
multi-metric, cross-technique comparison: we jointly evaluate privacy (recall), semantic
and factual utility (sentiment, topic stability, Q\&A and Q\&A-true), and trainability
under privacy via LoRA fine-tuning, across multiple anonymization methods (DLP, Presidio,
ZSTS in two variants, and local LLM-based substitution). This unified comparative
framework quantifies the privacy--utility--trainability trade-off and establishes
operational thresholds within a Responsible-AI-by-Design paradigm, thereby transforming
fragmented evidence into actionable guidance for LLMOps / AgentOps over sensitive text
data.

A complementary strand of research seeks to operationalize Responsible AI (RAI) through
design patterns, metrics, guardrails, and end-to-end lifecycle processes. Pattern
catalogues and governance playbooks~\cite{lu2024responsible} translate high-level
principles into concrete guidance spanning organizational, process, and product layers.
In parallel, RAI metrics catalogues~\cite{xia2024towards} advocate for measurable
indicators that connect governance objectives to technical signals. In this paper, we
instantiate a metrics-driven evaluation consistent with the ``Responsible AI Metrics
Catalogue'' perspective: we jointly quantify (i)~privacy preservation, (ii)~factual
utility via Q\&A-agent accuracy, and (iii)~trainability under privacy using our proposed
LoRA-based mean absolute error (LoRA-MAE).

The emergence of LLM agents has further prompted taxonomies of runtime guardrails under
an AI-safety-by-design perspective~\cite{shamsujjoha2024towards} and AgentOps frameworks
for observability and artifact lineage~\cite{dong2024agentops}. Building on these
insights, our study evaluates the quality of Q\&A agent responses when an intermediate
anonymization layer precedes the answering LLM. Concretely, we insert an on-premise,
type-preserving substitution stage that sanitizes conversational inputs before any model
invocation, thereby avoiding exposure of sensitive information to third-party APIs while
enabling responsible agent operation. This design allows us to quantify downstream
utility under privacy constraints and to examine whether such a pre-processing layer can
preserve task performance without compromising confidentiality.

Sector-specific studies, for example, in financial~\cite{lu2023developing} and
health~\cite{mallardi2025towards}, underscore the necessity of privacy-by-design in
highly regulated environments. Complementary privacy frameworks for foundation model
systems~\cite{bi2025privacy} situate technical safeguards alongside process and
governance controls, reinforcing the value of anonymous-by-construction corpora for both
model development and inference.

\section{Conclusions}

This work advances the state of practice by delivering a systematic, multi-metric,
cross-technique evaluation that turns scattered findings into operational guidance. While
recent studies have explored local LLM-based anonymization, our contribution is to
benchmark it head-to-head against the widely used baselines (Google DLP, Microsoft
Presidio) and a strong learning-based method (ZSTS, with and without substitution) using
a unified protocol and comparable metrics. Concretely, we jointly assess privacy
(Recall), semantic and factual utility (Sentiment stability, Topic distance, Q\&A and
Q\&A-true), and trainability under privacy (LoRA fine-tuning). Under this framework,
local LLM substitution attains state-of-the-art privacy (Recall $= 0.99$), high factual
utility (Q\&A / Q\&A-true up to $0.95$ and $0.96$ respectively), and low trainability
loss (LoRA-MAE $= 0.038$), thereby meeting privacy, utility, and adaptation requirements
simultaneously. These results demonstrate that on-prem, substitution-based anonymization
yields text that is both responsible to use and operationally valuable for agentic
pipelines and for downstream model tuning, when evaluated against clear, auditable
thresholds.

From a Responsible-AI Engineering (RAIE) standpoint, our contributions are twofold:
\begin{itemize}
  \item a type-consistent, on-prem substitution pipeline that preserves discourse
    coherence while enforcing conversation-level consistency; and
  \item LoRA fine-tuning as a gateable acceptance criterion, explicitly linking privacy
    safeguards to model trainability and downstream utility. Together with explicitly
    declared non-functional requirements (NFRs), continuous-integration (CI) gates that
    enforce privacy and utility thresholds, and traceable artifacts (prompts, seeds,
    model versions, hashes), these elements translate high-level principles---privacy-by-design
    and AI-safety-by-design---into deployable software controls for sensitive
    conversational data.
\end{itemize}

However, LLM anonymization tools have higher computational cost: local LLMs with 7B to
20B parameters require substantially more resources than rule-based alternatives
(DLP/NER) or lightweight masked language models. Nonetheless, organizations with existing
on-premise GPU infrastructure may find this overhead acceptable for production
deployments where privacy requirements prohibit external API usage. Future work will be
centred on reducing the cost via distillation, quantization, and batching/caching, and
evaluating smaller local models to approach parity with lightweight baselines. Also,
future work will broaden validation to multiple corpora, domains, and languages, and
extend trainability evaluation to classification, extraction, and summarization.

A second limitation of this work concerns dataset scope and evaluation scale: while ABCD
is a standard dialogue corpus in the literature, the Q\&A-based evaluation is conducted
on a random subset of 50 conversations. Although consistent with prior
work~\cite{albanese2023text}, this relatively small sample size may limit statistical
power and should be considered a potential threat to validity. Future work will expand
this analysis to larger subsets and additional datasets.

Third, trainability assessment: we evaluate a single downstream task (sentiment
regression) with one model configuration (BERT with LoRA) as an initial proxy. Future
research will address these constraints through several directions. Computational
efficiency can be improved via model distillation, quantization, batching optimizations,
and evaluation of smaller parameter models. Dataset coverage will expand through
validation across multiple corpora representing diverse conversational scenarios.
Trainability assessment will broaden to additional downstream tasks---classification,
extraction, summarization, question answering, retrieval augmented generation, and
agentic workflows---across different model architectures, establishing whether fine-tuning
metrics reliably indicate utility preservation across settings.

In summary, results show that substitution-based anonymization with local LLMs, coupled
with LoRA fine-tuning as a utility metric, can simultaneously safeguard privacy, preserve
utility, and maintain trainability. The results indicate that responsibly sanitized text
remains fit for agentic pipelines and downstream fine-tuning without unacceptable
degradation, underscoring the approach's promise for Responsible-AI-by-Design deployments
in real-world settings.


\end{document}